\documentclass[]{llncs}
\usepackage[numbers]{natbib}
\usepackage{graphicx}
\usepackage{amsmath,amssymb} 
\usepackage{color}
\usepackage{commath}
\usepackage{wrapfig}
\usepackage[width=122mm,left=12mm,paperwidth=146mm,height=193mm,top=12mm,paperheight=217mm]{geometry}
\usepackage{comment}
\graphicspath{{figs/}}

\begin{document}

\title{Virtual CNN Branching: Efficient Feature Ensemble for Person Re-Identification} %

\author{Albert Gong$^1$, Qiang Qiu$^2$\thanks{qiang.qiu@duke.edu}, Guillermo Sapiro$^2$}
\institute{$^1$North Carolina School of Science and Mathematics,  \\ 
$^2$Duke University
 }

\maketitle

\begin{abstract}
In this paper we introduce an ensemble method for convolutional neural network (CNN), called ``virtual branching,'' which can be implemented with nearly no additional parameters and computation on top of standard CNNs. We propose our method in the context of person re-identification (re-ID). Our CNN model consists of shared bottom layers, followed by ``virtual'' branches, where neurons from a block of regular convolutional and fully-connected layers are partitioned into multiple sets. Each virtual branch is trained with different data to specialize in different aspects, e.g., a specific body region or pose orientation. In this way, robust ensemble representations are obtained against human body misalignment, deformations, or variations in viewing angles, at nearly no any additional cost. The proposed method achieves competitive performance on multiple person re-ID benchmark datasets, including Market-1501, CUHK03, and DukeMTMC-reID. 
\end{abstract}

\section{Introduction}

\begin{figure}[t]
\centering
\includegraphics[height=9cm]{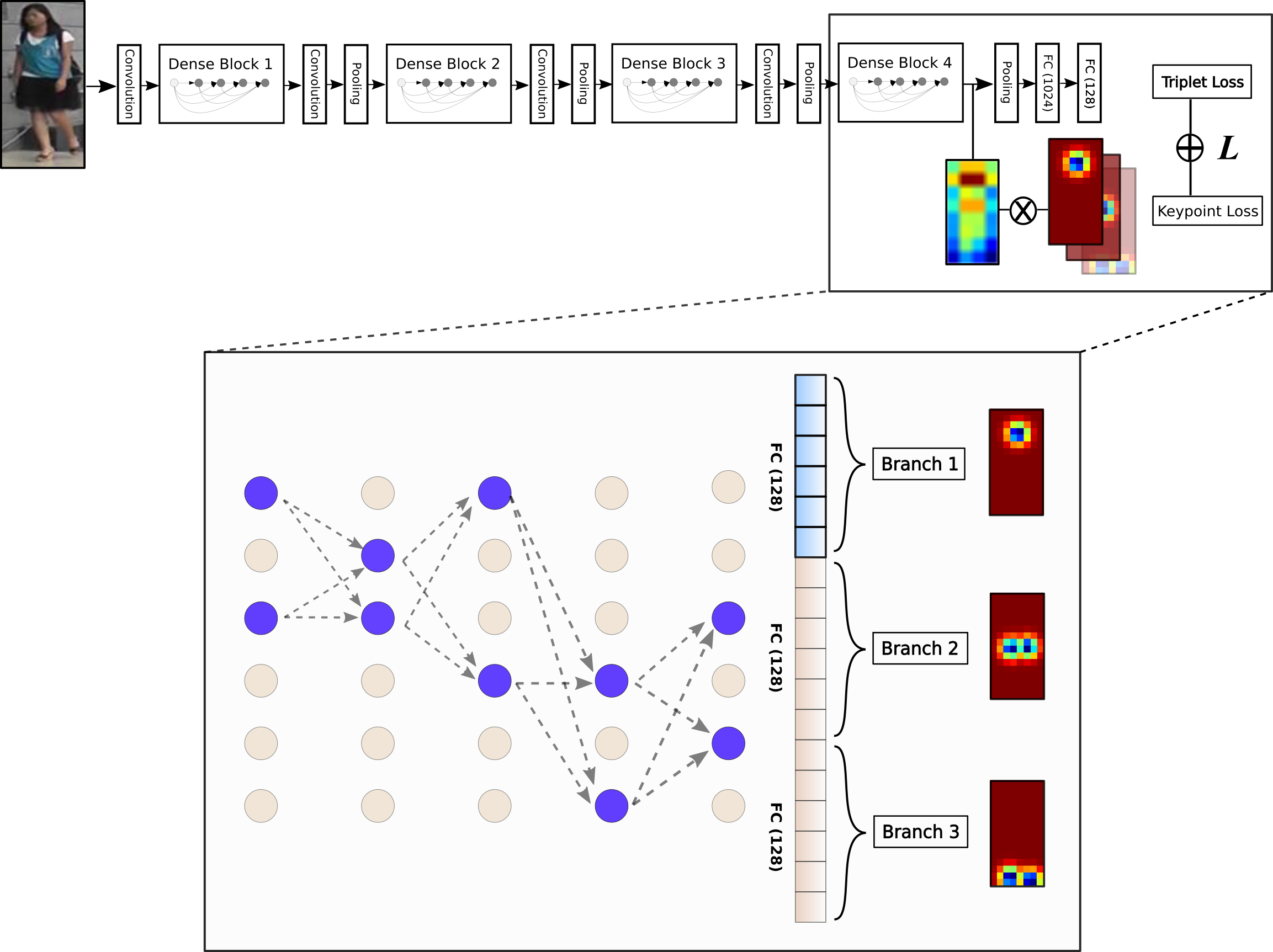}
\caption{Proposed model architecture. We adopt the DenseNet architecture with four dense blocks. We share parameters for the first three blocks, then partition neurons in the last block into virtual branches (purple neurons represent neurons that are selected for the first branch). With this architecture,  no additional parameters are needed. Each branch is learned using different data to specialize in different aspects (Section \ref{sec:localization}), e.g., according to human landmark information (shown) or pose orientation information. Triplet loss is used as the baseline loss function (Section \ref{sec:triplet}). For the example training scheme as shown, we also use the keypoint loss, a.k.a the localization-inducing loss component (Section \ref{sec:localization}).}
\label{fig:architecture}
\end{figure}

	In person re-identification (re-ID), the main goal is to find in a gallery set a matching image with the same identity as a query set.     
    This is a challenging problem due to the potentially large difference between the query and the matching image, such difference resulting from intrinsic changes, e.g., body position, to extrinsic, e.g., viewing angles. 
    Although a complete person re-ID system also involves person detection and person tracking, we use the term person re-ID primarily to refer to the task of person retrieval, which remains an open and heavily studied topic among the computer vision community \cite{Zheng2016}. In terms of real-world applications, automated person re-ID has great potential, especially within the fields of security, enabling efficient identification of human subjects from images or video data without human intervention.
     While motivated by this important problem, the overall framework here proposed is expected to have numerous applications, as it will be clear after fully describing it.

\begin{figure}[h]
\includegraphics[height=4cm]{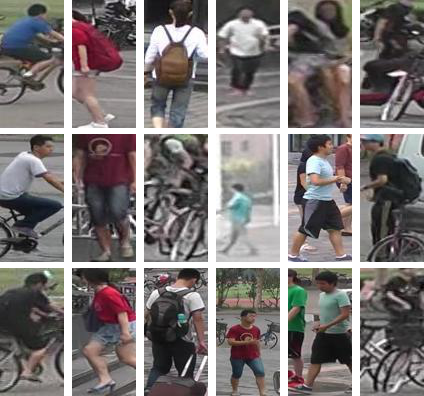}
\centering
\caption{Example challenges in person re-identification: body misalignment, pose deformation, missing parts, different viewing angles, and occlusion, etc. Images taken from Market-1501 dataset.}
\label{fig:market1501}
\end{figure}

	We propose a CNN ensemble architecture in Fig. \ref{fig:architecture} for person re-ID, to improve robustness to human body misalignments, deformations, or orientation variance (Fig. \ref{fig:market1501}). By training several models, each specialized for a different aspect, e.g., a specific body region or pose orientation, we construct a final ensemble whose predictions are more robust against body misalignment, pose deformations or variations in camera angles. Conventional approaches to ensemble deep learning are limited by the computational expense of training multiple deep models and longer inference times during testing. Thus, to make our proposed system usable,  an efficient method for implementing ensemble networks is needed that minimizes the number of additional parameters and that can be easily parallelized with limited computational resources, e.g., low GPU memory.
    
To enable our deep model to exhibit ensemble behavior without adding extra parameters, we introduce the method of ``virtual branching.'' Since lower-level features are often similar between deep convolutional neural networks, we construct a pseudo-ensemble network by inducing variation only in the top layers of the model, while sharing the low-level features in the bottom layers. Whereas in conventional ensemble learning, each member of the ensemble is often treated as a stand-alone model with information only being shared during the final prediction stage, virtual branching assigns a certain subset of the neurons in the original CNN model to each branch partition. Although no physical connections between neurons are broken, we omit activations of neurons not assigned to the same branch (Fig. \ref{fig:architecture}). Compared to Dropout \cite{Srivastava2014Dropout:Overfitting}, which is used primarily as a regularization technique, our proposed method fixes the neurons within each branch, and trains each set of neurons using different training data, allowing neurons to develop more specialized features. 

We illustrate the capability of the proposed ensemble architecture with two example training schemes, tailored to  the person re-ID task. In the first scheme, we extract keypoint locations of select human landmarks, e.g., shoulders, hips, and ankles, to generate heatmaps, which serve as labels to localize each branch to a specific body region, to improve robustness to body misalignments.
In the second scheme, we train each branch using subsets of the original dataset comprised of images of people with different pose orientations to improve robustness to different viewing angles. Compared to other person re-ID methods, e.g., \cite{Zheng2017}, \cite{Su2017}, and \cite{Zhao2017}, our method does not use a region proposal network and can easily be implemented on existing deep CNN models at little additional computation cost. Our method achieves state-of-the-art performance comparable to current methods on the Market-1501, CUHK03, and Duke-MTMC person re-ID datasets. 

The main contributions of this paper can be summarized as follows:     
\begin{itemize}
\item We propose an ensemble deep learning method, called ``virtual branching,'' that allows a network to exhibit ensemble behavior without additional parameters. Thus, our method does not incur significantly increased computation during both training and testing.
\item We introduce a localization-inducing loss function component that results in improved alignment of deep features around specific body regions for improved person re-ID performance.
\item We demonstrate that the proposed method can also produce deep features that are more robust to different pose orientations, i.e., variations in viewing angles.
\end{itemize}

\section{Related Work}

\subsection{Model Averaging}

Conventional ensemble methods, such as bagging \cite{Breiman1996BaggingPredictors} and boosting \cite{Friedman20019991}, have been shown to improve generalization performance of machine learning methods by combining several diverse models into a single predictor. 
 These early works clearly motivate the design of ensemble CNNs.
Recently, Dropout \cite{Srivastava2014Dropout:Overfitting} was proposed as a regularization technique to randomly omit neurons in the network, effectively resulting in an exponential number of different neural networks that are averaged into a single deep model. 
 While dropout is one of the responsible techniques for the current success of CNNs, the overall motivation and (semantic) structure is very different than the model here introduced.
Opitz \textit{et al.} \cite{Opitz} proposed a loss function to encourage diversity between individual networks during training for improved ensemble learning. Van der Laan \textit{et al.} \cite{VanDerLaanStatisticalLearner} proposed a cross-validation ensemble framework for creating a weighted combination of many candidate learners to build a ``super learner,'' which Ju \textit{et al.} \cite{JuTheClassification} demonstrated to outperform many conventional ensemble methods, i.e., unweighted averaging, majority voting. 
 As mentioned before, these methods, which were not developed for re-ID,  result in increased cost and lack the semantic tailored architecture here proposed.

\subsection{Person Re-Identification}
The two main approaches to deeply learned person re-ID algorithms have been (1) to treat person re-ID as a metric learning problem, and (2) to treat person re-ID as a classification problem. Before the release of large-scale person re-ID datasets, e.g., Market-1501 \cite{Zheng2015}, CUHK03 \cite{Li2014}, training of deeply learned models was limited to small datasets, e.g., VIPeR \cite{Gray2007}, i-LIDS \cite{Wei-Shi2011}, each consisting of only a few hundred training images. Thus, instead of a directly-trained identity classifier, a Siamese CNN was adopted to learn the similarity between pairs of images \cite{Li2014}, \cite{Yi2014}. This approach was expanded to triplets using the triplet loss \cite{Zhao2017}, \cite{Cheng2016}, \cite{Hermans2017}, \cite{Khamis2015}, \cite{Schroff2015}.

Recent research has combined a metric-based identification approach with a classification approach. Geng \textit{et al.} \cite{Geng2016} designed a model consisting of a person re-ID verification subnet that learns a softmax classifier with cross-entropy loss and a pairwise verification subnet that learns whether an input image pair contain the same person. Lin \textit{et al.} \cite{Lin2017} proposed a CNN that learns a re-ID embedding and predicts pedestrian attributes simultaneously using a combined classification loss function.

A common theme among person re-ID research has been to design methods that consider local features, in addition to global features, for improved invariance to confounding variables, such as pose distortion. Li \textit{et al.} \cite{Li2017} proposed a Joint Learning Multi-Loss CNN model that divides a global feature map into stripes, which are each then convolved further for more localized representations. Zheng \textit{et al.} \cite{Zheng2017}, \cite{Zheng2017a} applied affine transformations to the input image in order to produce more normalized pedestrian representations. Zhao \textit{et al.} \cite{Zhao2017} used a region proposal network as well as tree-structured feature fusion model to incorporate information about the human body. Wang \textit{et al.} \cite{Wang2017} focused on the vehicle re-ID task, and uses output response maps from a region proposal module to produce orientation invariant feature embeddings.
As we will show later in this paper, the simple ensemble method here introduced outperforms all these approaches without increasing the cost of existing CNNs.

\section{Methodology}

In this section, we illustrate in detail the virtual branching architecture, to enable ensemble behavior without adding extra parameters to the baseline CNN model. 
To demonstrate its capability to allow virtual branches being trained differently,
we then present two example training schemes for the person re-ID task. 
In the first scheme, we use human landmarks to localize each virtual branch to a specific body region, to address challenges such as body misalignments, deformation, and occlusion, etc. In the second scheme, we train each branch using subsets with different pose orientations to improve robustness to different viewing angles.

\subsection{Virtual Branching}

To construct our model, we assign neurons in the baseline CNN to specific partitions, which we call ``branches.'' 
Let $B_{i}$ be the set of neurons in the $i$-th branch.  
Each neuron represents a convolutional filter in a convolutional layer, and a fully-connected unit in a fully-connected layer.
Further, let $N^{l}=\{n_{1}^{l}, n_{2}^{l}...n_{k}^{l}\}$ be the set of $k$ neurons in the $l$-th layer of the model, and $L^{l}_{i} \subseteq N^{l}$ be the subset of $N^{l}$ in $B_{i}$. Thus,
\begin{equation*}
B_{i} = L^{1}_{i} \cup L^{2}_{i} \cup L^{3}_{i} \cup ... \cup L^{m}_{i}.
\end{equation*}
Let $b$ be the number of ``branches" in the model. Thus, each layer in the network (except the input layer) to which ``virtual branching'' is applied has $b$ outputs. During both testing and training, we multiply the $j$-th output of neuron $n_{i}^{l}$ in the $l$-th layer by the constant, $c$:
\begin{equation*}
c=\begin{cases}
1 , n_{i}^{l} \in B_{j}, \\
0 , otherwise,
\end{cases}
\end{equation*}
for all $1 \leq j \leq b$. To select the neurons for each branch, we define the hyperparameter, 
\begin{equation} \label{eq:1}
\delta = \frac{\sigma^{l}}{\sigma^{l} + \omega^{l}},
\end{equation}
to represent the degree of parameter sharing, where $\sigma^{l}$ is the number of neurons in the $l$-th layer that is common among all branches and $\omega^{l}$ is the number of neurons in the $l$-th layer that is unique to each branch. Let $d^{l}$ be the number of total neurons in the $l$-th layer, i.e., last dimension of the output of the $l$-th layer. Thus,
\begin{equation} \label{eq:2}
\sigma^{l} = \Big[\frac{d^{l}}{1 - b(1 - \frac{1}{\delta})}\Big],
\end{equation}
\begin{equation} \label{eq:3}
\omega^{l} = \Big[(\frac{1}{\delta} - 1)\sigma^{l}\Big],
\end{equation}
when $\delta > 0$; and $\sigma^{l} = 0$, $\omega^{l} = \frac{d^{l}}{b}$, when $\delta = 0$. In general, we set $L^{l}_{i}$ to include the neurons $n^{l}_{p}$, where $p \in [1,\sigma^{l}] \cup [\sigma^{l} + (i - 1)\omega^{l} , \sigma^{l} + i\omega^{l}]$, regardless of pre-initialized weights. 

\subsubsection{Architecture}
We initialize our model using the DenseNet architecture \cite{Huang2016} pre-trained on the ImageNet dataset \cite{JiaDeng2009}. Since person re-ID datasets are relatively small, only containing several thousands of samples, models trained on larger datasets, such as ImageNet \cite{JiaDeng2009}, contain knowledge that may be common to more specific domains or tasks. As done by Hermans \textit{et al.} \cite{Hermans2017}, we discard the final classification layer and add two fully-connected (FC) layers for our task; the first FC layer having 1024 units, followed by a batch normalization layer \cite{Ioffe2015} with ReLU activation \cite{Glorot2011}; the second FC layer having 128 units and also serving as the final feature embedding. 

Specifically for the DenseNet archiecture, we apply a drop mask layer after every convolutional and fully-connected layer in the last block \cite{Huang2016}. In image classification, a common challenge is determining the relative weights of predictions from multiple models. By treating person re-ID as a metric learning problem, we simply concatenate the embeddings from each branch after discarding the outputs from masked neurons to form the final embedding. 

\subsection{Human Landmark Localization}
\label{sec:localization}

\begin{figure}
\centering
\includegraphics[height=4cm]{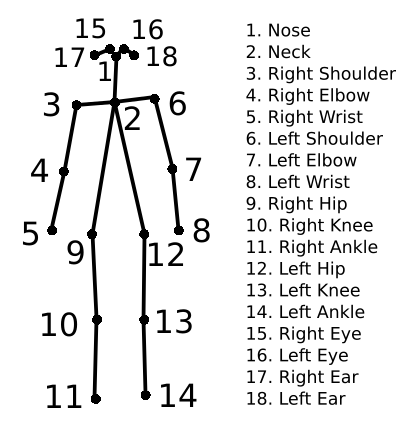} \hspace{0.5cm}
\includegraphics[height=4cm]{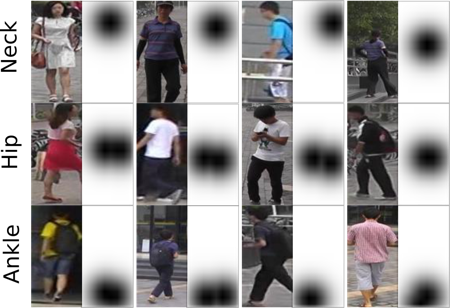}
\caption{Left: location of human keypoints \cite{Cao2016}; right: example heatmaps for neck, hip, and ankle body regions (Eq. \ref{eq:heatmap}).}
\label{fig:keypoints}
\end{figure}

We localize the features of each branch around specific human body regions in order to improve feature robustness to challenges such as body misalignments. deformation, and occlusion.
More specifically, we train each branch as a regressor in addition to the baseline identification task. Thus, this implementation essentially serves as a multi-task learning scenario. We generate separate heatmaps for the neck and hip regions and use the heatmaps as labels during training. Without loss of generality, let $k \in \mathbb{R}^{H\times W}$ represent the location of some given keypoint. Thus, we define the corresponding  heatmap, $H_{k}$ as
\begin{equation}
H_{k}(x,y) \triangleq 1 - e^{-\frac{(x-k_{x})^{2}-(y-k_{y})^{2}}{\sigma^{2}}}, \forall (x,y) \in \mathbb{R}^{H\times W},
\label{eq:heatmap}
\end{equation}
where $\sigma > 0$. For the hip and ankle regions, we take the element-wise maximum of the heatmaps for the right and left body parts in order to reduce the number of branches for training.

\subsubsection{Localization-inducing Loss Component}

In order to localize the feature embeddings of each branch around certain human body regions, we penalize activations produced by the layer just before the final global average pooling layer according to the respective body region heatmaps. Specifically, let $\alpha \in \mathbb{R}^{h \times w \times c}$ be the output of the final layer before the global average pooling operation for an input image, $i$, where $h$ is the output height, $w$ is the output width, and $c$ is the number of output “channels.” In order to calculate the value of the localization-inducing loss component, $L_{k}$, we first compute the channel-wise average, $\alpha_{i,avg} \in \mathbb{R}^{h \times w}$:
\begin{equation*}
\alpha_{i,avg_{x,y}} = \frac{1}{c} \sum_{\gamma=1}^{c} \alpha_{i_{x,y,\gamma}}.
\end{equation*}
We then normalize $\alpha_{i,avg}$ such that $\alpha_{i,avg_{x,y}} \in [0,1]$,
\begin{equation*}
\beta_{i_{x,y}} = \alpha_{i,avg_{x,y}} - \min_{\substack{{p=1...w}\\{q=1...h}}} \alpha_{i,avg_{p,q}},
\end{equation*}
\begin{equation*}
\alpha_{i,norm_{x,y}}=\frac{\alpha_{i,avg_{x,y}}}{\max_{\substack{{p=1...w}\\{q=1...h}}} \beta_{i_{p,q}}}.
\end{equation*}

Without loss of generality, let $H_{i} \in [0,1]^{h_{k} \times w_{k}}$ (Eq. \ref{eq:heatmap}) be the corresponding keypoint heatmap for the input image, $i$, for some body region, $k$, i.e., neck, hips, or ankle. Note that the dimensions of the heatmap, i.e., $h_{k} \times w_{k}$, are not necessarily equal to the dimensions of the output activation map, $\alpha_{i}$, in which case bilinear interpolation is used to scale the activation to the dimensions of the heatmap. However, for the purposes of this paper, we use $h_{k} \times w_{k} = 8 \times 4$. Next, we define the localization-inducing loss function component: 
\begin{equation*}
L_{k}(\theta;\mathbf{X};H) \triangleq \sum_{i=1}^{PK} \sum_{x=1}^{w} \sum_{y=1}^{h} (\alpha_{i,norm_{x,y}} H_{i_{x,y}}),
\end{equation*}
where $\theta$ is some set of model parameters and $\mathbf{X}$ is the training batch. The final loss function is then a weighted sum of the baseline triplet loss, $L_{T}$ (Eq. \ref{eq:triplet}), and the localization-inducing loss component:
\begin{equation}
L_{k}(\theta;\mathbf{X};H) \triangleq L_{T}(\theta;\mathbf{X}) + \lambda L_{k}(\theta;\mathbf{X};H),
\end{equation}
where $\lambda > 0$ is some hyperparameter constant. For all experiments with the localization-inducing loss function component, we set $\lambda = 0.2$ to have an appropriate balance between training of the baseline identification and body-region localization tasks.

\subsection{Pose Orientation}
\label{sec:pose-orientation}

\begin{figure}
\centering
\includegraphics[height=5cm]{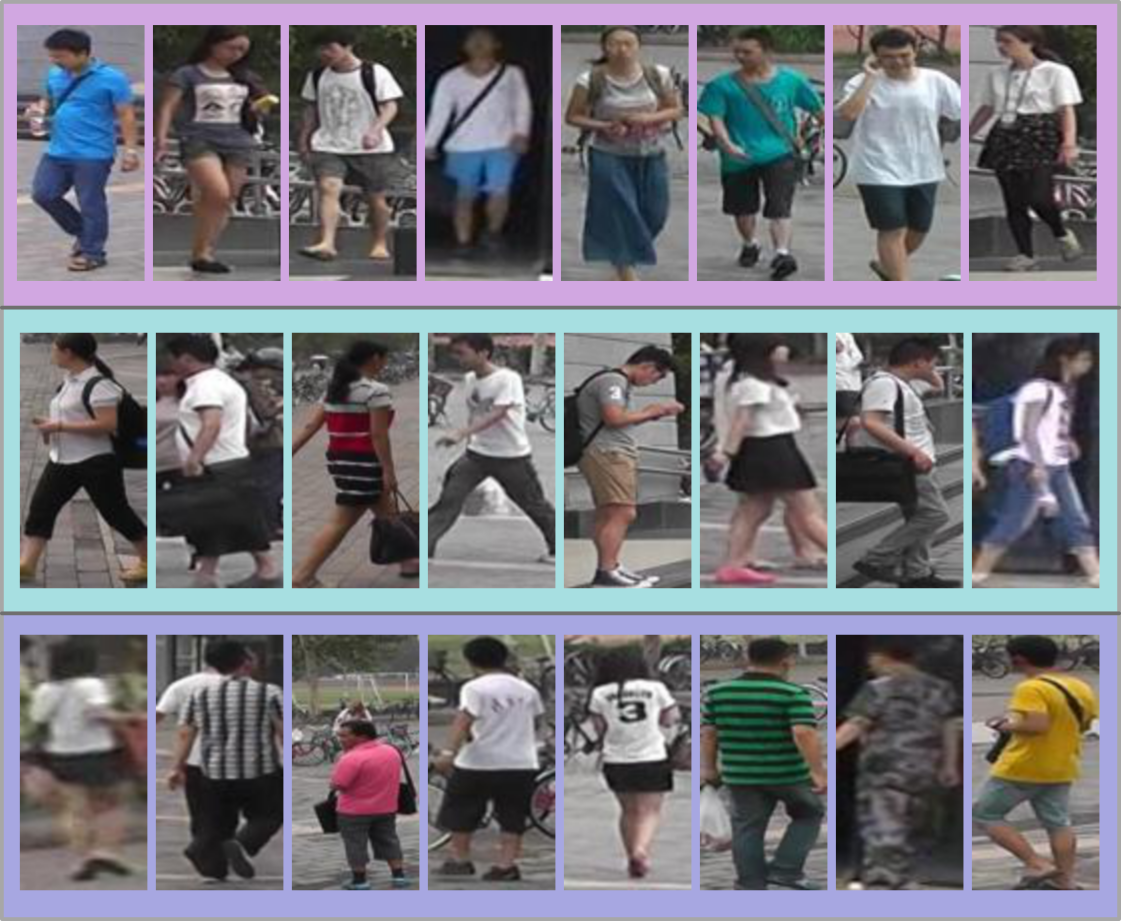}
\caption{Samples of people from front, side, and back views (top to bottom)  from the Market-1501 dataset.}
\label{fig:orientation}
\end{figure}

We also train each branch according to specific pose orientations, i.e., viewing angles, whereby each branch is trained on disjoint subsets of the training data, comprised of images of people taken from frontal, side, and back views (Fig. \ref{fig:orientation}). Essentially, we treat the training scenario as a multi-domain learning task, i.e., each pose orientation is a distinct domain. Thus, our model has three inputs, one for each data subset. We then train each branch using the baseline triplet loss (Section \ref{sec:triplet}).

To determine the orientation of a person within an image, we extract human keypoint information using a pose estimation algorithm, e.g., OpenPose \cite{Cao2016}. Specifically, we select the locations, $k$ of the right shoulder (RS), left shoulder (LS), right hip (RH), and left hip (LH): $k_{RS}, k_{LS}, k_{RH}, k_{LH} \in \mathbb{R}^{2}$ such that $(0,0)$ represents the top-left corner of an image. We then determine the angle of rotation as:
$\theta \approx cos^{-1}\Big(\frac{\mu D(k_{RS}, k_{LS})}{\frac{1}{2}(D(k_{RS}, k_{RH})+D(k_{LS}, k_{LH}))}\Big)$,
where $D(a,b)$ is the 2D Euclidean distance and $\mu = \frac{k_{RS,x}-k_{LS,x}}{\abs{k_{RS,x}-k_{LS,x}}}$, used to distinguish between images of people facing forward and people facing backward. Each image with orientation angle $\theta$ is assigned to its respective training subset $s$, 
\begin{equation*}
s(\theta)=\begin{cases}
front , 0 \leq \theta \leq \frac{\pi}{3}, \\
side , \frac{\pi}{3} \leq \theta \leq \frac{2\pi}{3}, \\
back , \frac{2\pi}{3} \leq \theta \leq \pi.
\end{cases}
\end{equation*}

\subsection{Baseline Loss}

\label{sec:triplet}

For all of the models, we use a triplet loss baseline, which aims to “pull together” input images from matching identities while “pushing away” input images from different identities. Thus, we learn a mapping $f_{\theta}(x):\mathbb{R}^{A}\rightarrow \mathbb{R}^{B}$, for some set of parameters $\theta$, such that semantically similar points in the input domain $\mathbb{R}^{A}$, are metrically close in the embedding vector space $\mathbb{R}^{B}$, and analogously semantically different points in $\mathbb{R}^{A}$ are metrically distant in $\mathbb{R}^{B}$.

We employ the batch-hard variant of the triplet loss \cite{Hermans2017}, which results in triplets that are more informative than trivially-chosen triplets. To construct a training batch, we randomly select $P$ person identities, then randomly sample $K$ images from each identity. We calculate the triplet loss, $L_{T}$, by selecting the hardest positives, $\rho$, and the hardest negatives, $\nu$, within each batch, $\textbf{X}$:
\begin{equation}
L_{T}(\theta;\mathbf{X}) \triangleq \sum_{i=1}^{P}\sum_{a=1}^{K}[m+\rho-\nu]_{+},
\label{eq:triplet}
\end{equation}
\begin{equation*}
\rho \triangleq \max_{p=1...K}D(f_{\theta}(x_{a}^{i}),f_{\theta}(x_{p}^{i})),
\end{equation*}
\begin{equation*}
\nu \triangleq \min_{\substack{{j=1...K}\\{j=1...K}\\{j\neq i}}}D(f_{\theta}(x_{a}^{i}),f_{\theta}(x_{p}^{i})),
\end{equation*}
where $\mathbf{x}_{a}$ is the anchor point, $\mathbf{x}_{p}$ is the positive point, and $\mathbf{x}_{n}$ is the negative point. We use the $L_1$ norm as the distance measure, $D$, and $[\bullet]_{+}=max(\bullet,0)$ as the hinge function, for some constant $m>0$. For all experiments, we use $m=0.2$ due to its good performance \cite{Hermans2017}.

\section{Experiment Setup}

\subsection{Datasets}

To evaluate our proposed method, we select Market-1501 \cite{Zheng2015}, CUHK03 \cite{Li2014}, and DukeMTMC-reID \cite{Ristani2016}, three large-scale person re-ID datasets, each comprised of images taken by an array of multiple cameras at university campuses. The large scale and use of automatic detection (for Market-1501 and CUHK03), which introduces increased image misalignment, allow these datasets to be more realistic compared to smaller person re-ID datasets.

\subsubsection{Market-1501.} This dataset contains 32,668 images of 1,501 identities. We use the provided training/testing split with 12,936 training images across 750 identities and 19,732 testing images across 751 identities. Following the evaluation protocol specified by Zheng \textit{et al.} \cite{Zheng2015}, we calculate the cumulative match score (CMC) and mean average precision (mAP) metrics to evaluate the performance of our models.

\subsubsection{CUHK03.} This dataset contains 14,097 images of 1,467 identities. For evaluation, the dataset provides 20 test sets, each with 100 identities. For each set, we use the 100 identities for testing and the rest 1,367 identities for training. We report the averaged performance after repeating the experiments for 20 times. We follow standard evaluation protocols to calculate CMC and mAP scores. This evaluation process is repeated for 100 times and the mean value is reported as the final result \cite{Li2014}.

\subsubsection{DukeMTMC-reID.} This dataset is a subset of Duke-MTMC \cite{Ristani2016} for image-based re-ID in the format of the Market-1501 dataset and is comprised of 36,411 images across 1,812 different identities from eight camera angles, where 1,404 identities appear in more than two cameras and the other 408 identities are considered distractor IDs. Among the 1,404 identities, 16,522 images of 702 identities are used for training, and the other 702 identities are divided into 2,228 query images and 17,661 gallery images.

\subsection{Training Implementation}
\label{sec:training}

All of our models are trained using the Keras deep learning library (version 2.0.8) with TensorFlow \cite{Abadi2016} (version 1.3.0) backend on a single NVIDIA K80 GPU (12 GB). Due to memory constraints, we set $P=18$ and $K=4$ when generating training data batches. Additionally, we use the Adam optimizer \cite{Kingma2014} with $lr_{0}=0.0003$, $\beta_{1}=0.9$, $\beta_{2}=0.999$, and $\varepsilon=1.0\times 10^{-8}$.

\subsubsection{Data Pre-processing.}
We find that simply resizing the input images yields best performance. 
For transfer learning on pre-trained ImageNet models, we noticed that using an input resolution similar to that of the pre-trained model yields the best results, likely because the pre-trained features are optimized for a certain scale. Thus, we set input images to $256 \times 128$ resolution for Market-1501 and DukeMTMC-reID and $256 \times 96$ resolution for CUHK03.

\subsubsection{Learning Rate Schedule.}
To train our models, we use a decaying learning rate:
$\epsilon(t)=\begin{cases}
\epsilon_{0} \\
\epsilon_{0} (\frac{1}{2})^{\lfloor \frac{t-t_{0}}{10} \rfloor}
\end{cases}$ at epoch $t$. We noticed that using the floor function yields the best performance by stabilizing the training. We train our models for 150 epochs, setting $t_{0}=50$, with 100 steps per epoch.

\subsection{Testing Implementation}

Following common practice \cite{Hermans2017, Krizhevsky2012}, we average the embeddings from horizontal flips of the original input image, yielding noticeable performance gains at the cost of acceptable increases in computation. As a distance measure between embeddings, we employ the Euclidean distance. After initial inference, the computations of the CMC and mAP metrics can easily be parallelized for reduced testing times.

\section{Experimental Results}

\begin{table}[b!]
\centering
\begin{tabular}{| c | c | c| c |c|c|c|}
\hline
 & \multicolumn{3}{|c|}{Single Query} & \multicolumn{3}{|c|}{Multi-query} \\
\hline
Method & mAP & Rank-1 & Rank-5 & mAP & Rank-1 & Rank-5 \\
\hline
Gated Siamese CNN \cite{Varior2016} & 39.95 & 65.88 & - & 48.45 & 76.04 & - \\
\hline
PIE \cite{Zheng2017} & 53.87 & 78.65 & 90.26 & - & - & - \\
\hline
JLML \cite{Li2017} & 65.5 & 85.1 & - & 74.5 & 89.7 & - \\
\hline
PAN \cite{Zheng2017} & 63.35 & 82.81 & - & 71.72 & 88.18 & - \\
\hline
Res50 + Attribute \cite{Lin2017} & 64.67 & 84.29 & 93.20 & - & - & - \\
\hline
GoogLeNet + DTL \cite{Geng2016} & 65.5 & 83.7 & - & 73.08 & 89.6 & - \\
\hline
PDC \cite{Su2017} & 63.41 & 84.14 & 92.73 & - & - & - \\
\hline
SpindleNet \cite{Zhao2017} & - & 76.9 & 91.5 & - & - & - \\
\hline
TriNet$\dagger$ \cite{Hermans2017} & 69.14 & 84.92 & 94.21 & 76.42 & 90.53 & 96.29 \\
\hline
MobileNet + DML$\dagger$ \cite{Zhang2017} & 68.86 & 87.73 & - & 77.14 & 91.66 & - \\
\hline
\multicolumn{7}{|c|}{} \\
\hline
\textbf{Baseline} & 68.3 & 83.3 & 93.1 & 75.8 & 90.1 & 95.9 \\
\hline
\textbf{Human Landmark} & 70.8 & \textbf{87.9} & 96.0 & \textbf{79.4} & \textbf{93.5} & \textbf{98.9} \\
\hline
\textbf{Pose Orientation} & \textbf{71.1} & 87.7 & \textbf{96.5} & 79.3 & 93.3 & 98.5 \\
\hline
\end{tabular}

\caption{Performance on Market-1501. Best in bold. $\dagger$: published in Arxiv preprint.}
\label{table:1}
\end{table}

\begin{table}[t!]
\centering
\begin{tabular}{| c | c | c| c |c|c|c|}
\hline
 & \multicolumn{3}{|c|}{Labeled} & \multicolumn{3}{|c|}{Detected} \\
\hline
Method & mAP & Rank-1 & Rank-5 & mAP & Rank-1 & Rank-5 \\
\hline
Gated Siamese CNN \cite{Varior2016} & - & - & - & 51.25 & 61.8 & 80.9 \\
\hline
PIE \cite{Zheng2017} & - & - & - & 67.21 & 61.50 & 89.30 \\
\hline
JLML \cite{Li2017} & - & 83.2 & 98.0 & - & 80.6 & 96.9 \\
\hline
PAN \cite{Zheng2017} & 35.03 & 36.86 & 56.86 & 34.00 & 36.29 & 55.50 \\
\hline
GoogLeNet + DTL \cite{Geng2016} & - & 85.4 & - & - & 84.1 & - \\
\hline
PDC \cite{Su2017} & - & 88.70 & - & - & 78.29 & - \\
\hline
SpindleNet \cite{Zhao2017} & - & - & - & - & 88.5 & 97.8 \\
\hline
TriNet$\dagger$ \cite{Hermans2017} & - & 89.63 & 99.01 & - & 87.58 & 98.17 \\
\hline
\multicolumn{7}{|c|}{} \\
\hline
\textbf{Baseline} & 93.6 & 88.2 & 99.4 & 92.0 & 86.4 & 98.1\\
\hline
\textbf{Human Landmark} & 95.5 & \textbf{91.0} & \textbf{99.8} & \textbf{94.7} & 88.9 & \textbf{99.4} \\
\hline
\textbf{Pose Orientation} & \textbf{95.8} & 90.9 & 99.6 & 94.5 & \textbf{89.3} & \textbf{99.4}\\
\hline
\end{tabular}

\caption{Performance on CUHK03. Best in bold. $\dagger$: published in Arxiv preprint.}
\label{table:2}
\end{table}

\begin{table}[t!]
\centering
\begin{tabular}{| c | c | c| c |}
\hline
Method & mAP & Rank-1 & Rank-5 \\
\hline
JLML \cite{Li2017} & 56.4 & 73.3 & - \\
\hline
PAN \cite{Zheng2017} & 51.51 & 71.59 & - \\
\hline
Res50 + Attribute \cite{Lin2017} & 51.88 & 70.69 & - \\
\hline
BoW + KissMe \cite{Liao2015} & 12.17 & 25.13 & - \\
\hline
LOMO + XQDA \cite{Wang2016} & 17.04 & 30.75 & - \\
\hline
Res50 + LSRO \cite{Zheng2017b} & 47.1 & 67.7 & - \\
\hline
SVDNet + Res50 \cite{SunSVDNetRetrieval} & 56.8 & 76.7 & - \\
\hline
\multicolumn{4}{|c|}{} \\
\hline
\textbf{Baseline} & 55.6 & 74.0 & 83.5 \\
\hline
\textbf{Human Landmark} & \textbf{58.4} & \textbf{76.1} & \textbf{85.7} \\
\hline
\textbf{Pose Orientation} & 58.2 & 75.7 & 84.3 \\
\hline
\end{tabular}

\caption{Performance on DukeMTMC-ReID. Best in bold.}
\label{table:3}
\end{table}

Performance comparisons on person re-ID  benchmarks are reported in tables \ref{table:1}, \ref{table:2}, and \ref{table:3}.
Our baseline model already outperforms many of the other previously published
methods,  due in part  to the use of the batch-hard variant of the triplet loss \cite{Hermans2017}. Although our baseline achieves marginally worse performance compared to TriNet \cite{Hermans2017}, which uses the same triplet loss function, our model uses significantly less parameters (8.2M) due to the use of the DenseNet architecture \cite{Huang2016}, compared to TriNet (25.8M parameters), which uses the ResNet-50 architecture \cite{He2016}. 

We immediately compensate for this slight performance drop using our proposed virtual branching method, which incorporates the Human Landmark and the Pose Orientation as introduced before. On average, we report a 3-4\% improvement over the baseline model on the Market-1501, CUHK03, and DukeMTMC-reID datasets. Our method exceeds or achieves comparable performance with current state-of-the-art methods on these datasets in terms of mAP and CMC rank-1 and rank-5 score.

We see that training each virtual branch using human landmark information achieves similar performance gains compared to training each virtual branch using pose orientation information. 
These results support our claim that combining several specialized models results in better overall performance. Furthermore, our proposed method provides an alternative to previous pose alignment methods, e.g., \cite{Zheng2017,Su2017,Zhao2017}, using region proposal networks or input-level transformations.

\subsection{Branching}

We compare models with varying numbers of branches and degrees of parameter sharing within each layer, defined by the hyperparameter $\delta$ (Eq. \ref{eq:1}). Only varying the number of branches, we notice that the performance increases significantly with the addition of several branches, but quickly experiences diminished returns since the number of neurons in each layer for each branch decreases as the number of branches increases (eqs. \ref{eq:2} and \ref{eq:3}). However, even with significantly less neurons than in the baseline model for the last few layers (i.e., 20-30\%), each branch is still able extract informative deep features. Additionally, the performance tends to increase as the degree of parameter overlap decreases. One possible explanation is that increasing the level of parameter sharing decreases the same level of induced variation that improves ensemble performance.

\begin{figure}
\centering
\includegraphics[height=5cm]{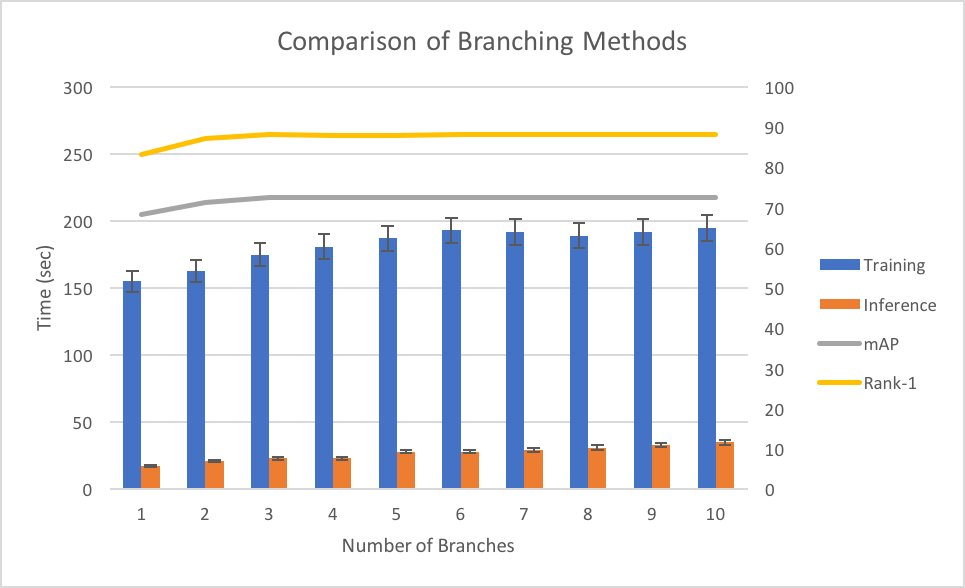}
\caption{Comparison of training and inference times and performance for models with different number of branches: training (seconds per epoch), inference (seconds per 1,000 samples), mAP (Market-1501), Rank-1 (Market-1501). See Section \ref{sec:training} for implementation details. }
\label{fig:times}
\end{figure}

We analyze the effects of our ensemble method on training and inference times (Fig. \ref{fig:times}). Theoretically, we should only see modest increases in computation times; although the use of branching should increase the number of operations per forward pass, branching only occurs in the top few layers and involves relatively few branches.

\subsection{Localization} 

\begin{figure}[h!]
\centering
\includegraphics[height=3cm]{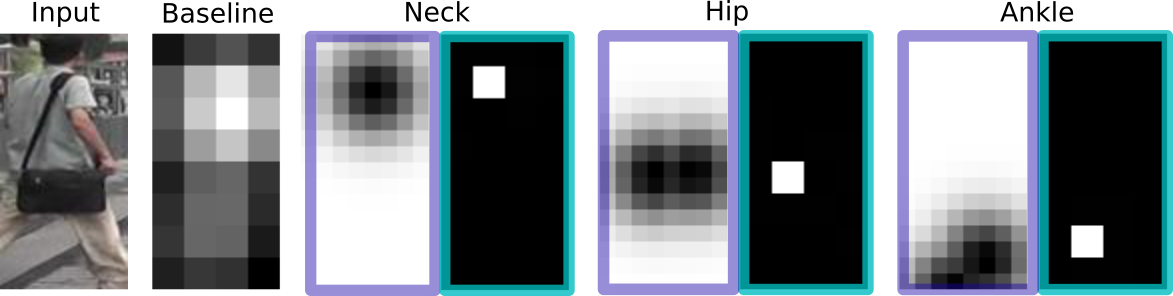}
\caption{Left to right: input image; user-generated activation map of baseline model before global average pooling layer (Section \ref{sec:pose-orientation}); for each of neck-localized model, hip-localized model, ankle-localized model: heatmap label (left) and learned activation map (right). }
\label{fig:regressor}
\end{figure}

We verify that our proposed method indeed encourages the models to localize around their corresponding human body regions. Recall that the localization-inducing loss function component penalizes against activations that are far away from the central keypoints (Section \ref{sec:localization}); we therefore expect that the activation map to which the localization-inducing loss is applied will have high response around the corresponding body region and low response elsewhere. Looking at the normalized mean output activations (defined in Section \ref{sec:localization}), we see that the activations indeed are more concentrated around the corresponding body regions, Fig. \ref{fig:regressor}.

\subsection{Feature Robustness}

\begin{figure}
\centering
\includegraphics[height=3cm]{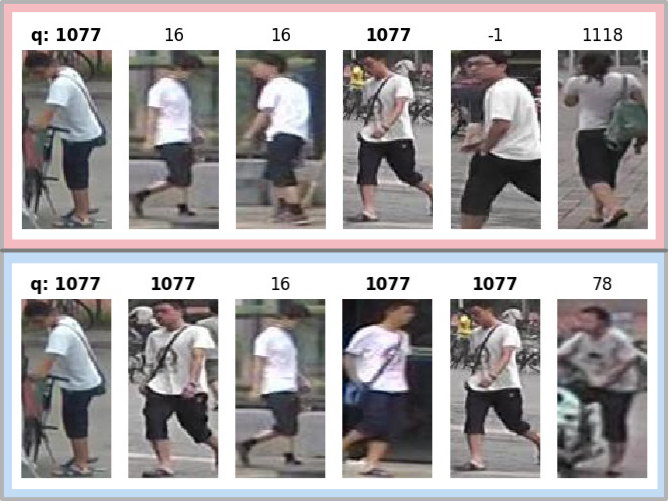}
\includegraphics[height=3cm]{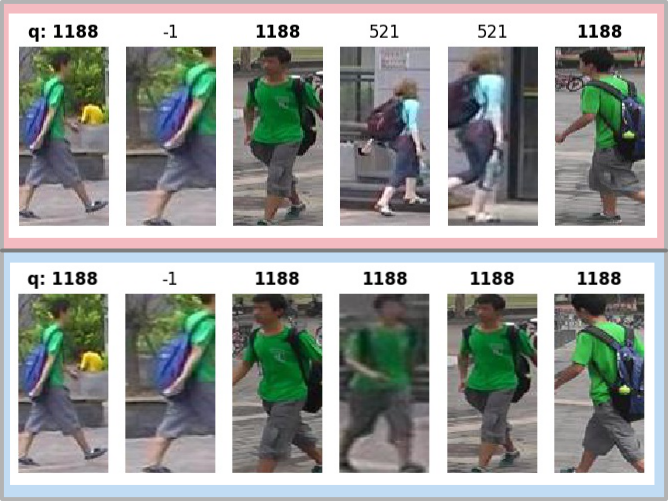}
\includegraphics[height=3cm]{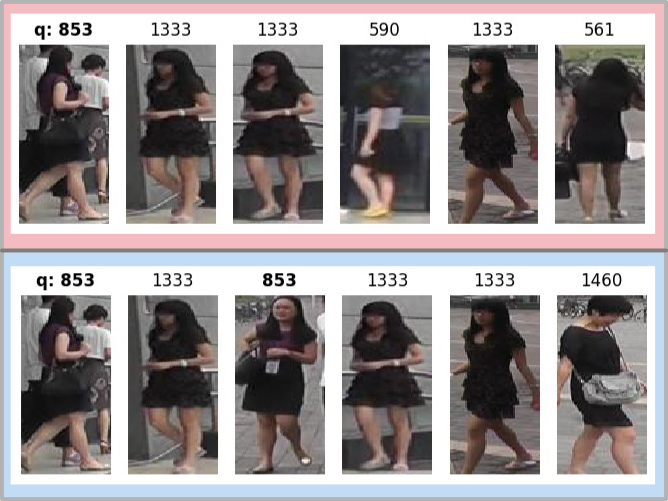}
\caption{Rankings examples of image retrieval  (query is the first image in each group). Top: rankings from baseline model; bottom: rankings from model using proposed method. Images are from the Market-1501 dataset.}
\label{fig:alignment}
\end{figure}

We also examine the robustness effect of the proposed ensemble features with landmark localized branches.
Looking at examples of rankings for image retrieval (Fig. \ref{fig:alignment}), we see that the proposed branching method indeed helps the deep model to become more robust to variability  in misalignment, pose deformation, occlusions, and other expected diversity in the re-ID challenges. Overall, these improvements lead to better performance on the entire testing set, in terms of mAP and rank-1 (tables \ref{table:1}, \ref{table:2}, \ref{table:3}).

\section{Conclusions}

In this paper, we proposed ``virtual branching,'' an ensemble method for person re-ID. The novel architecture  enables the deep model to exhibit ensemble behavior without significantly increased computation during training and testing. Specifically for the task of person re-ID, we train each branch according to different body regions and pose orientations and achieve state-of-the-art performance on several benchmark person re-ID datasets, i.e., Market-1501, CUHK03, and DukeMTMC-reID.
 The application of the proposed framework to other datasets and tasks is currently under investigation.

\section*{Acknowledgments}

Work partially supported by NSF, NIH, ONR, NGA, and ARO.

\bibliographystyle{splncs}
\bibliography{Mendeley}
\end{document}